\documentclass{article}

\usepackage{microtype}
\usepackage{graphicx}
\usepackage{subfigure}
\usepackage{booktabs} 

\usepackage{hyperref}

\usepackage[accepted]{icml2018_workshop}

\icmltitlerunning{Pooling of Causal Models under Counterfactual Fairness via Causal Judgement Aggregation}

\usepackage{array}
\usepackage{amsmath}
\usepackage{amsbsy}
\usepackage{amssymb}
\usepackage{float}
\usepackage{listings}
\usepackage{url}
\usepackage{hyperref}
\usepackage{todonotes}
\usepackage{tikz}

\graphicspath{{./img/}}

\newcommand{\TODO}[1]{}

\newcommand{\weight}[1]{w_{#1}}
\newcommand{\Nexpert}{N}
\newcommand{\oexpert}[1]{o_{#1}}
\newcommand{\opool}{o^*}
\newcommand{\pexpert}[2]{p_{#1} \left(#2\right)}
\newcommand{\ppool}[1]{p^* \left(#1\right)}
\newcommand{\jexpert}[2]{j_{#1} \left(#2\right)}
\newcommand{\jpool}[1]{j^* \left(#1\right)}
\newcommand{\Mexpert}[1]{\mathcal{M}_{#1}}
\newcommand{\Mpool}{\mathcal{M}^*}

\newcommand{\jarule}[1]{\texttt{JARule}\left( #1 \right)}

\newcommand{\setremove}{ \backslash }
\newcommand{\setunion}{ \cup }

\newcommand{\exvarset}[1]{\mathcal{U}_{#1}}
\newcommand{\envarset}[1]{\mathcal{V}_{#1}}
\newcommand{\structeqset}[1]{\mathcal{F}_{#1}}

\newcommand{\model}[1]{\mathcal{M}_{#1}}
\newcommand{\modeldiagram}[1]{\mathcal{G}\left({#1}\right)}
\newcommand{\modeldef}{\left( \exvarset{}, \envarset{}, \structeqset{} \right)}
\newcommand{\probmodel}[1]{\mathcal{M}_{#1}}
\newcommand{\probmodeldistr}{P\left(U\right)}
\newcommand{\probmodeldef}{\left( \exvarset{}, \envarset{}, \structeqset{}, \probmodeldistr \right)}

\newcommand{\parentVset}[1]{\mathcal{V}_{pa_{#1}}}
\newcommand{\parentUset}[1]{\mathcal{U}_{pa_{#1}}}
\newcommand{\parentV}[1]{v_{pa_{#1}}}
\newcommand{\parentU}[1]{u_{pa_{#1}}}
\newcommand{\descendent}[2]{Desc_{#2}\left(#1\right)}

\newcommand{\context}[1]{\overrightarrow{#1}}

\newcommand{\vertexset}[1]{\mathsf{V}_{#1}}
\newcommand{\edgeset}[1]{\mathsf{E}_{#1}}
\newcommand{\edge}[2]{{#1 \rightarrow #2}}
\newcommand{\vspool}[1]{\mathsf{V}_{#1}^*}
\newcommand{\espool}[1]{\mathsf{E}_{#1}^*}

\newcommand{\predY}{\hat{Y}}
\newcommand{\predYfunction}{f_{\predY}}
\newcommand{\protectedattribset}[1]{\mathcal{A}}
\newcommand{\featureset}[1]{\mathcal{X}}

\newtheorem{theorem}{Theorem}
\newtheorem{lemma}{Lemma}

\begin{document}
	
	\twocolumn[
	\icmltitle{Pooling of Causal Models under Counterfactual Fairness via Causal Judgement Aggregation}
	
	\begin{icmlauthorlist}
	\icmlauthor{Fabio Massimo Zennaro}{uio}
		\icmlauthor{Magdalena Ivanovska}{uio}
		%\icmlauthor{Audun J\o sang}{uio}
	\end{icmlauthorlist}

	\icmlaffiliation{uio}{Department of Informatics, University of Oslo, PO Box 1080 Blindern, 0316 Oslo, Norway}
	\setcounter{footnote}{1} 
	
	\icmlcorrespondingauthor{Fabio Massimo Zennaro}{fabiomz@ifi.uio.no}
	
	% You may provide any keywords that you
	% find helpful for describing your paper; these are used to populate
	% the "keywords" metadata in the PDF but will not be shown in the document
	\icmlkeywords{Machine Learning, ICML, Causality, Fairness, Opinion Pooling, Causal Judgement Aggregation, Counterfactual Fairness}
	
	\vskip 0.3in
	]
	
	% this must go after the closing bracket ] following \twocolumn[ ...
	
	% This command actually creates the footnote in the first column
	% listing the affiliations and the copyright notice.
	% The command takes one argument, which is text to display at the start of the footnote.
	% The \icmlEqualContribution command is standard text for equal contribution.
	% Remove it (just {}) if you do not need this facility.
	
	\printAffiliations{}  % leave blank if no need to mention equal contribution
	%\printAffiliationsAndNotice{\icmlEqualContribution} % otherwise use the standard text.
	
	\begin{abstract}
		%Probabilistic causal models proved to be an expressive and rigorous language to formalize many real-world scenarios. In the case our models are representing sensitive scenarios, learning may be concerned not only with issues of accuracy, but also with issues of social fairness.
		In this paper we consider the problem of combining multiple \emph{probabilistic causal models}, provided by different experts, under the requirement that the aggregated model satisfies the criterion of \emph{counterfactual fairness}. We build upon the work on causal models and fairness in machine learning, and we express the problem of combining multiple models within the framework of \emph{opinion pooling}. We propose two simple algorithms, grounded in the theory of counterfactual fairness and causal judgment aggregation, that are guaranteed to generate aggregated probabilistic causal models respecting the criterion of fairness, and we compare their behaviors on a toy case study.   
	\end{abstract}

\TODO{his/her}
\TODO{improve algorithm}
\section{Introduction}

In this paper we try to bring together three different strands of research: \emph{causality}, \emph{fairness}, and \emph{opinion pooling}. 
Causality deals with definition and the study of causal relationships; \citet{pearl2009causality} provides a solid theoretical foundation for modelling causal structures. 
Fairness is concerned with guaranteeing that prediction systems deployed in sensitive scenarios support decisions that are fair from a social point of view; this topic is particularly relevant to the current research in machine learning, where models are often non-transparent and it is hard to evaluate whether their outputs are affected by discriminatory biases.
Opinion pooling tackles the challenge of aggregating the opinions of several experts; when these opinions are expressed as probability distribution functions (pdf) the problem of pooling is expressed as the problem of merging multiple pdfs in a single distribution that can be used by a decision maker.

The pair-wise intersection of these fields has been the object of recent research. 
\citet{kusner2017counterfactual} and \citet{kilbertus2017avoiding} analyzed how concepts from the domain of causality (counterfactuals, unresolved discrimination, proxy discrimination) may be used to assess fairness in the context of causal graphs.
\citet{bradley2014aggregating} and \citet{alrajeh2018combining} offered a first attempt at extending the problem of opinion pooling over causal models and discussed how to aggregate potentially incompatible causal graphs.

However, little research has been done so far on the problem of aggregating causal models given a requirement of fairness. \citet{russell2017worlds} propose that the problem of learning a classifier under multiple fairness constraints derived from causal models provided by different experts can be recast as an optimization problem through an $\epsilon$-relaxation in the definition of fairness.

In this paper we offer a first exploration of the issue of combining expressive and realistic models under the requirement of fairness using the conceptual framework of opinion pooling. 
In particular, building upon the work on aggregating causal judgments in \citet{dietrich2016probabilistic}, we analyze the case in which potentially unfair models must be pooled and we show how results from the work on counterfactual fairness and causal judgment aggregation may be used to generate aggregated fair models. We design two complementary algorithms for producing aggregated fair models and we explore their properties.
%We will first show that, under this framework, the aggregation of fair models is guaranteed to return an aggregated fair model. 

The rest of the paper is organized as follows: Section \ref{sec:Background} reviews basic concepts in the research areas considered; Section \ref{sec:Aggregation} provides the formalization of our problem and our contribution; Section \ref{sec:Conclusion} draws conclusions on this work and indicates future avenues of research.

\section{Background \label{sec:Background}}

This section reviews basic notions used to define the problem of aggregating probabilistic causal models under fairness: Section \ref{ssec:Causality} recalls the primary definitions in the study of causality; Section \ref{ssec:Fairness} discusses the notion of fairness in machine learning; Section \ref{ssec:OpPooling} offers a formalization of the problem of opinion pooling.

\subsection{Causality \label{ssec:Causality}}
Following the formalism in \citet{pearl2009causality}, we provide the basic definitions for working with causality\footnote{For a more complete treatment of these concepts, refer to \citet{pearl2009causality}.}.  

\paragraph{Causal Model}
We define a \emph{causal model} $\model{}$ as a triple $\modeldef$ where:
\begin{itemize}
	\item $\exvarset{}$ is a set of exogenous variables $\left\lbrace U_1, U_2, ..., U_m \right\rbrace$ representing background factors that are not affected by other variables in the model;
	\item $\envarset{}$ is a set of endogenous variables $\left\lbrace V_1, V_2, ..., V_n \right\rbrace$ representing factors that are affected by other variables in the model;
	\item $\structeqset{}$ is a set of structural equations $\left\lbrace f_1, f_2, ..., f_n \right\rbrace$ such that the value $v_i$ of variable $V_i$ is $v_i = f_i \left( \parentV{i}, \parentU{i} \right)$, where $\parentV{i}$ are the values assumed by the variables in the set $\parentVset{i} \subseteq \envarset{}\setremove V_i$ and $\parentU{i}$ are the values assumed by the variables in the set $\parentUset{i} \subseteq \exvarset{}$; for each endogenous variable $V_i$, the structural equation $f_i$ defines the value $v_i$ as a function of a subset of variables in $\exvarset{} \setunion \envarset{}$.
\end{itemize}

\paragraph{Causal Diagram}
The \emph{causal diagram} $\modeldiagram{\model{}}$ associated with the causal model $\model{}$ is the directed graph $\left(\vertexset{},\edgeset{}\right)$ where:
\begin{itemize}
	\item $\vertexset{}$ is the set of vertices representing the variables $\exvarset{} \setunion \envarset{}$ in $\model{}$;
	\item $\edgeset{}$ is the set of edges determined by the structural equations in $\model{}$; edges are connecting each endogenous node $V_i$ with the set of its \emph{parent nodes} $\parentVset{i} \setunion \parentUset{i}$; we denote $\edge{V_j}{V_i}$ the edge going from $V_j$ to $V_i$. 
\end{itemize}
Assuming, by intuition, the acyclicity of causality, we will take that a causal model $\model{}$ entails an acyclic causal diagram $\modeldiagram{\model{}}$ represented as a \emph{directed acyclic graph} (DAG).

\paragraph{Context}
Given a causal model $\model{} = \modeldef$, we define \emph{context} $\context{u}$ as a specific instantiation of the exogenous variables, $U_1 = u_1, U_2 = u_2, ..., U_m = u_m$.\\
Given an endogenous variable $V_i$, we will use the shorthand notation $V_i\left(\context{u}\right)$ to denote the value of the variable $V_i$, given by propagating the context $\context{u}$ through the causal diagram according to the structural equations.

\paragraph{Intervention}
Given a causal model $\model{} = \modeldef$, we define \emph{intervention} $do(V_i = \bar{v})$ as the substitution of the structural equation $v_i = f_i \left( \parentV{i}, \parentU{i} \right)$ with the value $\bar{v}$.\\
Given two endogenous variables $X$ and $Y$, we will use the shorthand notation $Y_{X\leftarrow x}$ to denote the value of the variable $Y$ under the intervention $do(X = x)$.\\
Notice that, from the point of view of the causal diagram, performing the intervention $do(X = x)$ is equivalent to setting the value of the variable $X$ to $x$ and removing all the incoming edges $\edge{\cdot}{X}$ in $X$.

\paragraph{Counterfactual} 
Given a causal model $\model{} = \modeldef$, the context $\context{u}$, two endogenous variables $X$ and $Y$, and the intervention $do(X = x)$, we define \emph{counterfactual} the value of the expression $Y_{X\leftarrow x}(\context{u})$.\\
Conceptually, the counterfactual represents the value that $Y$ would have taken in context $\context{u}$ had the value of $X$ been $x$.

\paragraph{Probabilistic Causal Model}
A \emph{probabilistic causal model} $\probmodel{}$ is a tuple $\probmodeldef$ where:
\begin{itemize}
	\item $\modeldef$ is a causal model;
	\item $\probmodeldistr$ is a probability distribution over the exogenous variables. The probability distribution $\probmodeldistr$, combined with the dependence of each endogenous variable $V_i$ on the exogenous variables, as specified in the structural equation $f_i$, allows us to define a probability distribution $P(V)$ over the endogenous variables as: $P(V=v)=\sum_{\{ u \vert V = v \}} P(U= \context{u})$.
\end{itemize}

Notice that we overload the notation $\probmodel{}$ to denote both (generic) causal models and probabilistic causal models; the context will allow the reader to distinguish between them.

\subsection{Fairness \label{ssec:Fairness}}
Following the work of \citet{kusner2017counterfactual}, we review the topic of fairness, with a particular emphasis on counterfactual fairness for predictive models.

\paragraph{Fairness and Learned Models}
The study of \emph{fairness} is concerned with the societal impact of the adoption of machine learning models at large. When training machine learning systems on historical real-world data, the use of black-box models in sensitive contexts such as judicial sentencing or educational grants allocation, %or police enforcement 
carries the risk of perpetuating, or even introducing \citep{kusner2017counterfactual}, socially or morally unacceptable discriminatory biases 
(for a survey, see, for instance, \citealp{zliobaite2015survey}). 
Thus, fairness requires the definition of new metrics that take into account not only the performance of a machine learning system, but also its impact on actual and delicate situations in the real world. 

\paragraph{Fairness of a Predictor}
A predictive model can be represented as a (potentially probabilistic) function of the form $\predY = f(Z)$, where $\predY$ is a \emph{predictor} and $Z$ is a vector of covariates. According to an observational approach to fairness, the set of covariates is partitioned in a set of \emph{protected attributes} $\protectedattribset{}$, representing discriminatory elements of information, and a set of \emph{features} $\featureset{}$, carrying no overt sensitive information.
The predictive model is then redefined as $\predY = f(A,X)$ and the fairness problem is expressed as the problem of learning a predictor $\predY$ that does not discriminate with respect to the protected attributes $A$. 
Now, given the complexities of social reality and disagreement over what constitutes a fair policy, different measures and principles of fairness may be adopted to rule out discrimination. For instance, \emph{fairness through unawareness} is defined by requiring the predictor $\predY$ not to use protected attributes $A$ in its decision making; for a more thorough review of different types of fairness and their limitations, see \citet{gajane2017formalizing} and \citet{kusner2017counterfactual}.

\paragraph{Fairness Over Causal Models}
Given a probabilistic causal model $\probmodel{} = \probmodeldef$ fairness may be evaluated following an observational approach. Let us take $\predY$ to be an endogenous variable whose structural equation $\predYfunction$ is a predictive function; let us also partition the remaining endogenous variables $\envarset{} \setminus \{\predY\}$ into a set of protected attributes $\protectedattribset{}$ and a set of features $\featureset{}$. Then, a fair predictor $\predY$ is a function of the endogenous nodes, $\predY=\predYfunction \left( V \right) = \predYfunction\left( A,X \right)$, that respects a given definition of fairness. 

\paragraph{Counterfactual Fairness}
Given a probabilistic causal model $\probmodel{} = \probmodeldef$, a predictor $\predY$, and a partition of the endogenous variables $\envarset{} \setminus \{\predY\}$ into $\left(\protectedattribset{}, \featureset{} \right)$, the predictor $\predYfunction\left( A,X \right)$ is \emph{counterfactually fair} if, for every context $\context{u}$, $ P\left( \predY_{A\leftarrow a} (\context{u}) \vert X=x, A=a \right) =  
P\left( \predY_{A\leftarrow a'} (\context{u}) \vert X=x, A=a \right)$, for all values $y$ of the predictor, for all values $a'$ in the domain of $A$, and for all $x$ in the domain of $X$ \cite{kusner2017counterfactual}.\\
In other words, the predictor $\predY$ is counterfactually fair if, under all the contexts, the prediction given the protected attributes $A=a$ and the features $X=x$ would not change if we were to intervene $do(A=a')$ to force the value of the protected attributes $A$ to $a'$, for all the possible values that the protected attribute can assume.\\
Denoting $\descendent{\protectedattribset{}}{\model{}}$ the descendants of the nodes in $\protectedattribset{}$ in the model $\model{}$, an immediate property follows from the definition of counterfactual fairness:
\begin{lemma}\label{lemma}
	(Lemma 1 in \citealp{kusner2017counterfactual}) Given a probabilistic causal model $\probmodel{} = \probmodeldef$, a predictor $\predY$ and a partition of the endogenous variables $\left( \protectedattribset{}, \featureset{} \right)$, the predictor $\predY=\predYfunction\left( A,X \right)$ is counterfactually fair if $\predYfunction$ is a function only of variables that are not in $\descendent{\protectedattribset{}}{\model{}}$.
\end{lemma}

\subsection{Opinion Pooling \label{ssec:OpPooling}}
Following the study of \citet{dietrich2016probabilistic}, we introduce the framework for opinion pooling.

\paragraph{Pooling} Assume there are $\Nexpert$ experts, each one expressing his/her opinion $\oexpert{i}$, $1 \leq i \leq \Nexpert$. The problem of \emph{pooling} (or \emph{aggregating}) the opinions $\oexpert{i}$ consists in finding a single pooled opinion $\opool$ representing the merging of all the individual opinions.

\paragraph{Probabilistic Opinion Pooling}
Given opinions in the form of probability distributions $\pexpert{i}{x}$, \emph{probabilistic opinion pooling} means finding a single pooled probabilistic opinion of the form $\ppool{x} = f\left(\pexpert{i}{x}\right)$, where $f: [0,1]^{\Nexpert} \rightarrow [0,1]$ \cite{dietrich2016probabilistic}.\\
Now, given a set probabilistic opinions $\pexpert{i}{x}$, different functions $f$ may be chosen to perform opinion pooling, such as arithmetic averaging or geometric averaging. A grounded approach to choosing a function $f$ is based on the \emph{axiomatic approach}, that is, on the definition of a set of properties that the pooling function is required to satisfy \citep{dietrich2016probabilistic}. For instance, it can be shown that the only pooling function satisfying a property of event-wise independence (i.e., the value of $\ppool{x}$ depends only on the probability values assigned to $x$ by the $\Nexpert$ experts) and unanimity-preservation (i.e.: if all the $\Nexpert$ experts hold the same opinion $\pexpert{i}{x}=k$, then $\ppool{x}=k$) is the \emph{weighted linear pooling function} $\ppool{x} = \sum_{i=1}^{\Nexpert} \weight{i}\pexpert{i}{x}$, $0\leq\weight{i}\leq1$, $\sum_{i=1}^{\Nexpert} \weight{i} = 1$, where $\weight{i}$ is a weight assigned to the experts \cite{aczel1980characterization}; for a more in-depth review of different types of probabilistic opinion pooling functions and their properties, see \citet{dietrich2016probabilistic}. 

\paragraph{Judgement Aggregation}
Given opinions in the form of judgments, that is, true-or-false assignments $\jexpert{i}{x}$, \emph{judgement pooling} is concerned with defining a single pooled judgement of the form $\jpool{x} = f\left(\jexpert{i}{x}\right)$, where $f: \{0,1\}^{\Nexpert} \rightarrow \{0,1\}$ \cite{grossi2014judgment}.\\
As in the case of probabilistic opinion pooling, given a set of judgments $\jexpert{i}{x}$, different functions $f$ may be chosen to perform judgment pooling, such as majority voting or intersection. Again a grounded approach for choosing a function $f$ is an axiomatic approach \cite{bradley2014aggregating}.

\paragraph{Aggregation of Causal Models}

\citet{bradley2014aggregating} offer a seminal study of the problem of aggregating probabilistic causal models $\Mexpert{i}$ expressed as \emph{Bayesian networks}, that is structured representations of the joint probability distribution of a set of variables $\envarset{}$ and their conditional dependencies in the form of a DAG with associated conditional probability distributions \cite{pearl2014probabilistic}.
 
First, they show that a naive probabilistic opinion pooling over the conditional probabilities in the models $\Mexpert{i}$ is unable to preserve the basic property of conditional independence encoded in a Bayesian network, that is, even if for all the experts $\pexpert{i}{X,Y \vert Z} = \pexpert{i}{X \vert Z} \pexpert{i}{Y \vert Z}$, it can not be guaranteed that $\ppool{X,Y \vert Z} = \ppool{X \vert Z} \ppool{Y \vert Z}$.

They then suggest a \emph{two-stage approach} to the problem of aggregating the probabilistic causal models $\Mexpert{i}$ into a pooled probabilistic causal model $\Mpool$.
In the first qualitative step they reduce the problem of finding the graph associated with the pooled model $\modeldiagram{\Mpool}$ to the problem of judgment aggregation over the edges $\edge{X}{Y}$ in all the models $\Mexpert{i}$; %alternatively put, 
for all the possible edges $\edge{X}{Y}$ in all the models $\Mexpert{i}$, they take the presence of the edge from node $X$ to node $Y$ in model $\Mexpert{i}$ as the $i$-th expert casting the judgment $\jexpert{i}{\edge{X}{Y}}=1$, and the absence of it as the judgment $\jexpert{i}{\edge{X}{Y}}=0$; the problem of defining the diagram $\modeldiagram{\Mpool}$ is then tackled as a judgment aggregation problem over each edge $\edge{X}{Y}$.
In the second quantitative step, they reconstruct the joint probability of the pooled model $\Mpool$ solving a problem of probabilistic opinion pooling over the conditional distributions defining the joint of $\Mpool$.

A central result in the analysis of \citet{bradley2014aggregating} is the following impossibility result:
\begin{theorem} \label{theorem}
	(Theorem 1 in \citealp{bradley2014aggregating}) Given a set of probabilistic causal models $\Mexpert{i}$, if the set of vertices $\vertexset{i}$ contains three or more variables, then there is no judgment aggregation rule $f$ that satisfies all the following properties:
	\begin{itemize}
		\item \emph{Universal Domain}: the domain of $f$ includes all logically possible acyclic causal relations;
		\item \emph{Acyclicity}: the pooled causal model $\Mpool$ produced by $f$ is guaranteed to be acyclic;
		\item \emph{Unbiasedness}: given two vertices $X,Y \in \vertexset{i}$, the causal dependence of $X$ on $Y$ in $\Mpool$ rests only on whether $X$ is causally dependent on $Y$ in the individual $\Mexpert{i}$, and the aggregation rule is not biased towards a positive or negative outcome;
		\item \emph{Non-dictatorship}: the pooled causal model $\Mpool$ produced by $f$ is not trivially the causal model $\Mexpert{i}$ held by a given expert $i$.
	\end{itemize} 
\end{theorem} 
As a result of this theorem, no unique aggregation rule can be chosen for the pooling of causal judgments in the first step of the two-stage approach. A relaxation of these properties must be decided depending on the scenario at hand.

\section{Aggregation of Causal Models Under Fairness \label{sec:Aggregation}}

This section analyzes how probabilistic causal models can be aggregated under fairness: Section \ref{ssec:ProblemFormal} provides a formalization of our problem; Section \ref{ssec:Algorithms} presents two complementary algorithms for performing aggregation of probabilistic causal models under counterfactual fairness; finally, Section \ref{ssec:Illustration} offers an illustration of the use of our algorithms on a toy case study.
%\ref{ssec:Agg_Fair_Models} offers a simple result on the preservation of counterfactual fairness under causal judgment aggregation, while Section \ref{ssec:Agg_Unfair_Models} discusses proposes a greedy algorithm to aggregate counterfactually unfair models into a counterfactually fair model.

\subsection{Problem Formalization \label{ssec:ProblemFormal}}

We now consider the case in which $\Nexpert$ experts are required to provide a probabilistic causal model $\Mexpert{i} = \left( \exvarset{i}, \envarset{i}, \structeqset{i}, P_i(U) \right)$ representing a potentially sensitive scenario. We will make the simplifying assumption that the experts will provide the model $\Mexpert{i}$ over the same variables $\left(\exvarset{}, \envarset{}\right)$, so that the probabilistic causal model takes the form $\Mexpert{i} = \left( \exvarset{}, \envarset{}, \structeqset{i}, P_i(U) \right)$, where $\left(\exvarset{}, \envarset{}\right)$ are the nodes of the graph $\modeldiagram{\model{i}}$, $P_i(U)$ the probability distributions at the root nodes of the graph, and $\structeqset{i}$ the set of structural equations defining the behaviour of the remaining nodes.
To simplify the task of the expert, we will also take that the pdfs $P_i(U)$ and the functional form of the predictor $\predY = \predYfunction \left( V \right)$ are pre-defined; in particular, $\predYfunction$ is assumed to be a given algorithm (such as a neural network), while the inputs $V$ of this algorithm must be specified by the experts.\\ 
To sum up, the experts are required to specify a probabilistic causal model $\Mexpert{i}$ and to define which variables in the model are to be used in the predictor $\predY$. 

We also assume that the models provided by the experts are not necessarily fair, at least not in terms of counterfactual fairness as defined in Section \ref{ssec:Fairness}. We consider this assumption very reasonable as individual experts may not be aware of the potential for discrimination in their models or may not know how to formally guarantee fairness. 
The final decision maker, though, is aware of fairness implications and wants to generate an aggregated predictive probabilistic causal model $\Mpool$ that guarantees counterfactual fairness. As such, the decision maker is responsible for specifying the partitioning of the endogenous variables $\envarset{} \setminus \{\predY\}$ into $\left( \protectedattribset{}, \featureset{} \right)$; in other words, he/she is in charge of defining which variables are sensitive.

In conclusion, our problem may be expressed as: given $\Nexpert$ potentially counterfactually unfair probabilistic causal models $\Mexpert{i}$, a predictor $\predY$, and a partition of the variables $\envarset{} \setminus \{\predY\}$ into $\left( \protectedattribset{}, \featureset{} \right)$, is there a pooling function $f$ over causal models $\Mpool = f\left(\Mexpert{i}\right)$ such that $\Mpool$ is guaranteed to respect counterfactual fairness:
$ P_{\Mpool}\left( \predY_{A\leftarrow a} (\context{u}) \vert X=x, A=a \right) =  
P_{\Mpool}\left( \predY_{A\leftarrow a'} (\context{u}) \vert X=x, A=a \right)$?  

\subsection{Algorithms \label{ssec:Algorithms}}
To tackle our problem we adopt the two-stage approach for the aggregation of probabilistic causal models discussed in Section \ref{ssec:OpPooling}. Our solution focuses on the first step of this method: we modify the qualitative step in order to generate an aggregated graph $\modeldiagram{\Mpool}$ that guarantees counterfactual fairness, while we do not discuss the second quantitative step in which the structural equations are pooled.

There are two challenges in our approach: (i) we need the predictor $\predY$ in the aggregated probabilistic causal model $\Mpool$ to be indeed counterfactually fair; and (ii) we need to specify a relaxation of one of the intuitive requirements detailed in Theorem \ref{theorem}. The first challenge can be addressed by satisfying the condition in Lemma \ref{lemma}; to meet the second challenge we argue that the existence of a predictor $\predY$ in the graph $\modeldiagram{\Mpool}$ suggests the possibility of dropping the property of unbiasedness and provides a natural starting point for imposing an ordering on the edges of $\modeldiagram{\Mpool}$. 

Concretely, we tackle the two challenges above in the following two algorithmic steps:
\begin{itemize}
	\item \emph{Removal step}: remove all the protected attributes and their descendants from the aggregated model $\Mpool$;
	\item \emph{Pooling step}: perform judgment aggregation after ordering the edges in each graph $\modeldiagram{\model{i}}$ according to the distance from the predictor $\predY$.
\end{itemize}

The removal step enforces Lemma $\ref{lemma}$ and thus guarantees counterfactual fairness. 
The pooling step performs judgment aggregation relying on an ordering of the edges in each $\model{i}$ as a function of the distance from the predictor $\predY$. This ordering is not total, and we may still need a rule to break ties (e.g., random selection or an alphabetical criterion). Progressing according to this order, we can select edges and perform judgment aggregation using a concrete rule (e.g., majority or intersection). New edges are then added to the graph of the pooled model $\modeldiagram{\Mpool}$, as far as acyclicity is not violated \cite{bradley2014aggregating}.
Now, according to the order of these two steps, two different algorithms arise.

Algorithm $\ref{alg:filt-pool}$ reports the \emph{removal-pooling} algorithm, in which removal is performed first (lines 3-10) and then pooling (lines 12-28). Notice that this algorithm may have a high likelihood of producing an empty set of edges $\espool{}$ for the aggregated probabilistic causal model $\modeldiagram{\Mpool}$. This is due to the fact that in the removal stage we remove all the nodes that are descendant of the protected attributes $\protectedattribset{}$ in \emph{any} probabilistic causal model $\model{i}$. This reflects a very \emph{prudent} approach, in which even a single expert relating a protected attribute $A$ to a node $V_i$ is sufficient to remove all the nodes along paths starting in $A$ and going through $V_i$. This minimizes the risk of introducing in the final pooled probabilistic causal models $\Mpool$ variables that are potentially discriminatory and that were identified only by a single expert. On the other hand, the drawback of this approach is that few spurious connections from a protected attribute $A$ added by a unreliable expert may lead to an empty set of edges $\espool{}$.

\begin{algorithm} 
	\caption{Removal-Pooling Algorithm for Aggregation of Causal Models under Counterfactual Fairness}
	\label{alg:example}
	\begin{algorithmic}[1]\label{alg:filt-pool}
%		\STATE Build the graph backward from $\predY$ making it non-dependant on descendant of $A$
		\STATE {\bfseries Input:} $\Nexpert$ graphs models $\modeldiagram{\model{j}}$ over the variables $\envarset{}$, a predictor $\predY=\predYfunction(V)$ with different inputs over the $\Nexpert$ experts, a partitioning of the variables $\envarset{} \setminus \{\predY\}$ into protected attributes $\protectedattribset{}$ and $\featureset{}$, a judgment aggregation rule $\jarule{e}$  
%		\REPEAT
%		\IF{$x_i > x_{i+1}$}
%		\STATE $noChange = false$
%		\ENDIF
		\STATE
		\STATE Initialize $\envarset{fair} := \envarset{}$
		\FOR{$j=1$ {\bfseries to} $\Nexpert$}
			\STATE $\envarset{\neg} := \lbrace V \vert 
			\left(V \in \protectedattribset{}\right) \vee 
			\left( \in \descendent{\protectedattribset{}}{\model{j}}\right) \rbrace$
			\STATE $\envarset{fair} := \envarset{fair} \setminus \envarset{\neg}$
		\ENDFOR
		\FOR{$j=1$ {\bfseries to} $\Nexpert$}
			\STATE Remove from the edge set $\edgeset{j}$ of $\modeldiagram{\model{j}}$ all edges $\edge{V_x}{V_y} \vert \left(V_x \notin \envarset{fair} \vee V_y \notin \envarset{fair} \right)$
		\ENDFOR
		\STATE

		\STATE Initialize $D$ to the length of the longest path in the models $\model{j}$
		\STATE Initialize $\Mpool$ by setting up the graph $\modeldiagram{\Mpool}$ in which $\vspool{}:=\envarset{fair}$ and $\espool{}:=\emptyset$

		\FOR{$j=1$ {\bfseries to} $\Nexpert$}
			\STATE Initialize the vertex set $\vertexset{j,0}:={\predY}$
			\STATE Initialize the edge set $\edgeset{j,0}:=\emptyset$
		\ENDFOR
		
		\FOR{$j=1$ {\bfseries to} $\Nexpert$}
		\FOR{$d=1$ {\bfseries to} $D$}
			\STATE $\edgeset{j,d} := \lbrace \edge{V_x}{V_y} \vert \edge{V_x}{V_y} \in \edgeset{j} \wedge \left( V_x \in \vertexset{j,d-1} \vee V_y \in \vertexset{j,d-1}  \right)  \rbrace$
			
			\STATE $\vertexset{j,d} := \lbrace V_x \vert \left( \edge{V_x}{\cdot} \in \edgeset{j,d} \vee \edge{\cdot}{V_x} \in \edgeset{j,d}  \right) \rbrace$				
		\ENDFOR
		\ENDFOR
		
		\FOR{$j=1$ {\bfseries to} $\Nexpert$}
		\FOR{$d=1$ {\bfseries to} $D$}
		\STATE $\forall \left(\edge{V_x}{V_y}\right) \in \edgeset{j,d}$,\\
		if $\left( \jarule{\edge{V_x}{V_y}} =1 \right)$\\
		$\vee \left( \espool{} \setunion \{\edge{V_x}{V_y}\} \textnormal{ is acyclic} \right)$\\  
		then $\espool{} := \espool{} \setunion \{\edge{V_x}{V_y}\}$				
		\ENDFOR
		\ENDFOR

		\STATE

		\RETURN $\Mpool$

%		\UNTIL{$noChange$ is $true$}
	\end{algorithmic}
\end{algorithm}

Algorithm $\ref{alg:pool-filt}$ reports the \emph{pooling-removal} algorithm, in which pooling is performed first (lines 3-19) and then removal (lines 21-23). Differently from the previous algorithm, this solution is less likely to end with an empty set of edges $\espool{}$ for the aggregated probabilistic causal model $\modeldiagram{\Mpool}$. This is due to the fact that edges are first pooled and spurious connections introduced by unreliable experts are filtered out. This reflects a more \emph{compromise-based} approach, where some form of agreement (as defined by the concrete rule for judgment aggregation) is required to assert the causal influence of a protected attribute $A$ on a node $V_i$.

\begin{algorithm} 
	\caption{Pooling-Removal Algorithm for Aggregation of Causal Models under Counterfactual Fairness}
	\begin{algorithmic}[1] \label{alg:pool-filt}
		%		\STATE Build the graph backward from $\predY$ making it non-dependant on descendant of $A$
		\STATE {\bfseries Input:} $\Nexpert$ graphs models $\modeldiagram{\model{j}}$ over the variables $\envarset{}$, a predictor $\predY=\predYfunction(V)$ with different inputs over the $\Nexpert$ experts, a partitioning of the variables $\envarset{} \setminus \{\predY\}$ into protected attributes $\protectedattribset{}$ and $\featureset{}$, a judgment aggregation rule $\jarule{e}$
		%		\REPEAT
		%		\IF{$x_i > x_{i+1}$}
		%		\STATE $noChange = false$
		%		\ENDIF
		\STATE
		\STATE Initialize $D$ to the length of the longest path in the models $\model{j}$
		\STATE Initialize $\Mpool$ by setting up the graph $\modeldiagram{\Mpool}$ in which $\vspool{}=\envarset{}$ and $\espool{}=\emptyset$

		\FOR{$j=1$ {\bfseries to} $\Nexpert$}
		\STATE Initialize the vertex set $\vertexset{j,0}:={\predY}$
		\STATE Initialize the edge set $\edgeset{j,0}:=\emptyset$
		\ENDFOR
		
		\FOR{$j=1$ {\bfseries to} $\Nexpert$}
		\FOR{$d=1$ {\bfseries to} $D$}
		\STATE $\edgeset{j,d} := \lbrace \edge{V_x}{V_y} \vert \edge{V_x}{V_y} \in \edgeset{j} \wedge \left( V_x \in \vertexset{j,d-1} \vee V_y \in \vertexset{j,d-1}  \right)  \rbrace$
		
		\STATE $\vertexset{j,d} := \lbrace V_x \vert \left( \edge{V_x}{\cdot} \in \edgeset{j,d} \vee \edge{\cdot}{V_x} \in \edgeset{j,d}  \right) \rbrace$				
		\ENDFOR
		\ENDFOR
		
		\FOR{$j=1$ {\bfseries to} $\Nexpert$}
		\FOR{$d=1$ {\bfseries to} $D$}
		\STATE $\forall \left(\edge{V_x}{V_y}\right) \in \edgeset{j,d}$,\\
		if $\left( \jarule{\edge{V_x}{V_y}} =1 \right)$\\
		$\vee \left( \espool{} \setunion \{\edge{V_x}{V_y}\} \textnormal{ is acyclic} \right)$\\  
		then $\espool{} := \espool{} \setunion \{\edge{V_x}{V_y}\}$					
		\ENDFOR
		\ENDFOR		
		
		\STATE
		\STATE \STATE $\envarset{\neg} := \lbrace V \vert 
		\left(V \in \protectedattribset{}\right) \vee 
		\left( \in \descendent{\protectedattribset{}}{\model{j}}\right) \rbrace$
		\STATE $\vspool{} := \vspool{} \setminus \envarset{\neg}$
		\STATE Remove from the edge set $\espool{}$ of $\modeldiagram{\Mpool}$ all edges $\edge{V_x}{V_y} \vert \left(V_x \notin \vspool{} \vee V_y \notin \vspool{} \right)$
		
		\STATE
				
		\RETURN $\Mpool$
				
		%		\UNTIL{$noChange$ is $true$}
	\end{algorithmic}
\end{algorithm}

\subsection{Illustration \label{ssec:Illustration}}

Here we give a simple illustration of the problem of causal model aggregation under counterfactual fairness and we point out the effect and the differences between the removal-pooling algorithm and the pooling-removal algorithm.

The head of the Department of Computer Science has decided to develop a predictive model to help with the selection of PhD candidates. To do so, he/she has chosen to build a predictive model $f(Z)$ implemented as a neural network. In order to decide which features to use in $f(Z)$, Prof. Alice and Prof. Bob had been asked to define a causal model for this selection problem. Alice and Bob are both provided with the resumes of the candidates, from which they extract the following variables: \emph{age} (Age), \emph{gender} (Gnd), \emph{MSc university department} (Dpt), \emph{MSc final mark} (Mrk), \emph{years of job experience in Computer Science} (Job), \emph{quality of the cover letter} (Cvr), and the predictor ($\predY$). (For simplicity, we leave implicit the presence of an exogenous variable for each endogenous node).

Figures \ref{fig:M_A} and \ref{fig:M_B} illustrate $\modeldiagram{\model{A}}$ and $\modeldiagram{\model{B}}$, respectively, that is the diagrams associated with the causal model defined by Alice and Bob. The two graphs are very similar. Understandably, both agree that the decision on whether to admit a candidate or not should depend on his/her work experience in computer science, the department where he/she got his/her MSc degree, the MSc final mark, and the quality of the cover letter; they also agree that the amount of years of job experience in computer science is causally influenced by the age of the candidate and, since they both read reports on the gender gap in the computer science industry, they also think that gender affects the opportunity of the candidate of having a job.\\ 
On the other hand, the two models exhibit some differences. Specifically, after skimming through the study on gender bias in admissions at Berkley \citep{bickel1975sex,pearl2009causality}, Alice concludes that gender causally affects the choice of department where the candidate did his/her studies. Also, differently from Bob, she decides to exclude age as an input to the predictor.\\
The complete probabilistic causal model would require Alice and Bob to specify structural equations on the nodes, as well. However, since our algorithms work on the graphs only at a qualitative level, we omit the definition of these equations.\\ 
Notice that in their modeling Alice and Bob did not concern themselves with the issue of discrimination. %: their main objective was to develop a descriptive causal model that can be used for prediction. Notice 
Also notice that, from a purely formal point of view, they could have made the predictor depend on all the available variables; however, it seems more reasonable for a modeler to consider only those variables that are expected to affect the predictor.
%this is a possible and legitimate choice, even if it may seem more 

\begin{figure}
	\centering
	
	\begin{tikzpicture}[shorten >=1pt, auto, node distance=3cm, thick, scale=0.8, every node/.style={scale=0.8}]
	
	\tikzstyle{node_style} = [circle,draw=black]
	%\tikzstyle{edge_style} = [draw=black, line width=2, ultra thick]
	\node[node_style] (gnd) at (0,0) {Gnd};
	\node[node_style] (dpt) at (0,-2) {Dpt};
	\node[node_style] (job) at (2,0) {Job};
	\node[node_style] (mrk) at (2,-3) {Mrk};
	\node[node_style] (age) at (4,1) {Age};
	\node[node_style] (cvr) at (4,-3) {Cvr};
	\node[circle,draw=black,dashed] (y) at (6,-1) {$\predY$};
	
	\draw[->]  (gnd) edge (job);
	\draw[->]  (gnd) edge (dpt);
	\draw[->]  (dpt) edge (mrk);
	\draw[->]  (age) edge (job);
	
	\draw[->]  (cvr) edge (y);
	\draw[->]  (job) edge (y);
	\draw[->]  (dpt) edge (y);
	\draw[->]  (mrk) edge (y);
	
	\end{tikzpicture}
	
	\caption{Graph $\modeldiagram{\model{A}}$ of the causal model provided by Alice. \label{fig:M_A}}
\end{figure}

\begin{figure}
	\centering
	
	\begin{tikzpicture}[shorten >=1pt, auto, node distance=3cm, thick, scale=0.8, every node/.style={scale=0.8}]
	
	\tikzstyle{node_style} = [circle,draw=black]
	%\tikzstyle{edge_style} = [draw=black, line width=2, ultra thick]
	\node[node_style] (gnd) at (0,0) {Gnd};
	\node[node_style] (dpt) at (0,-2) {Dpt};
	\node[node_style] (job) at (2,0) {Job};
	\node[node_style] (mrk) at (2,-3) {Mrk};
	\node[node_style] (age) at (4,1) {Age};
	\node[node_style] (cvr) at (4,-3) {Cvr};
	\node[circle,draw=black,dashed] (y) at (6,-1) {$\predY$};
	
	\draw[->]  (gnd) edge (job);
	\draw[->]  (dpt) edge (mrk);
	\draw[->]  (age) edge (job);
	
	\draw[->]  (cvr) edge (y);
	\draw[->]  (job) edge (y);
	\draw[->]  (dpt) edge (y);
	\draw[->]  (mrk) edge (y);
	\draw[->]  (age) edge (y);
	
	\end{tikzpicture}
	
	\caption{Graph $\modeldiagram{\model{B}}$ of the causal model provided by Bob. \label{fig:M_B}}
\end{figure}

Now, the head of the department decides to aggregate the two models taking into account the policies for fairness approved by the University. Gender is marked as the only protected attribute, thus giving rise to the following partition of the endogenous nodes: $\protectedattribset{}=\{\textnormal{Gnd}\}$, $\featureset{}=\{\textnormal{Age},\textnormal{Dpt},\textnormal{Mrk},\textnormal{Job},\textnormal{Cvr}\}$. As a concrete judgment aggregation rule, the \emph{strict majority rule} is adopted, thus preserving edges in the pooled graph only when both Alice and Bob agree on the existence of a given edge.

Suppose now that the head of the department decides to use the \emph{removal-pooling algorithm}. In the first step of the algorithm, all the protected attributes and their descendants are removed. In Alice's model, $\{\textnormal{Gnd},\textnormal{Job},\textnormal{Dpt},\textnormal{Mrk}\}$ are removed, while in Bob's model only $\{\textnormal{Gnd},\textnormal{Job}\}$ are removed. At the end, we are left with sub-models of $\model{A}$ and $\model{B}$ illustrated in Figure \ref{fig:Filt_Pool}.
In the second step of the algorithm, the remaining nodes are ordered according to their distance from the predictor $\predY$ and pooled using the strict majority aggregation rule. This aggregation produce a minimal graph with the single feature $\{\textnormal{Cvr}\}$. The aggregated predictor, in order to be fair, is just $\predY=f\left( \textnormal{Cvr} \right)$, meaning that the decision should only be based on the quality of the cover letter.

\begin{figure}
	\centering
	
	\begin{tikzpicture}[shorten >=1pt, auto, node distance=3cm, thick, scale=0.8, every node/.style={scale=0.8}]
	
	\tikzstyle{node_style} = [circle,draw=black]
	%\tikzstyle{edge_style} = [draw=black, line width=2, ultra thick]
	\node[node_style] (age1) at (0,0) {Age};
	\node[node_style] (cvr1) at (0,-2) {Cvr};
	\node[circle,draw=black,dashed] (y1) at (2,-1) {$\predY$};
	
	\node[node_style] (age2) at (6,0) {Age};
	\node[node_style] (cvr2) at (7,-3) {Cvr};
	\node[circle,draw=black,dashed] (y2) at (8,-1) {$\predY$};	
	\node[node_style] (dpt2) at (4,-2) {Dpt};
	\node[node_style] (mrk2) at (5.5,-3) {Mrk};
			
	\draw[->]  (cvr1) edge (y1);
	
	\draw[->]  (age2) edge (y2);
	\draw[->]  (cvr2) edge (y2);
	\draw[->]  (dpt2) edge (y2);
	\draw[->]  (mrk2) edge (y2);
	\draw[->]  (dpt2) edge (mrk2);
	
	\end{tikzpicture}
	
	\caption{Graph of the sub-model $\modeldiagram{\model{A}}$ (left) and graph of the sub-model $\modeldiagram{\model{B}}$ (right) after the first step of the \emph{removal-pooling} algorithm. \label{fig:Filt_Pool}}
\end{figure} 

Let us suppose now that the head of the department decides to use the \emph{pooling-removal algorithm}. In the first step of the algorithm, edges in $\model{A}$ and $\model{B}$ are ordered according to the distance from the predictor (using an alphabetical criterion for resolution of ties):
\[
\begin{array}{cc}
d_{A,0}=\{\predY\} & d_{B,0}=\{\predY\}\\

d_{A,1}=\{\textrm{Cvr} \rightarrow \predY, & 
d_{B,1}=\{\textrm{Age} \rightarrow \predY, \\

\textrm{Dpt} \rightarrow \predY, \textrm{Job} \rightarrow \predY, & 
\textrm{Cvr} \rightarrow \predY, \textrm{Dpt} \rightarrow \predY\\

\textrm{Mrk} \rightarrow \predY \} & 
\textrm{Job} \rightarrow \predY, \textrm{Mrk} \rightarrow \predY \}\\

d_{A,2}=\{\textrm{Age} \rightarrow \textrm{Job}, & 
d_{B,2}=\{\textrm{Age} \rightarrow \textrm{Job}, \\

\textrm{Dpt} \rightarrow \textrm{Mrk}, \textrm{Gnd} \rightarrow \textrm{Dpt}, & 
\textrm{Dpt} \rightarrow \textrm{Mrk}, \textrm{Gnd} \rightarrow \textrm{Job} \}\\

\textrm{Gnd} \rightarrow \textrm{Job} \} &

\end{array}
\]
\normalsize
Following this ordering, edges are aggregated using the strict majority rule, giving rise to the pooled (not counterfactually fair) model $\Mpool$ in Figure \ref{fig:Pool_Filt}. 
In the second step of the algorithm, protected attributes and their descendants are removed from the pooled model $\Mpool$; that is, we remove $\{\textnormal{Gnd},\textnormal{Job}\}$. The final counterfactually fair predictor is $\predY=f\left( \textnormal{Dpt}, \textnormal{Mrk}, \textnormal{Cvr} \right)$, meaning that decisions can be taken on the basis of the department of the candidate, his/her final mark and his/her cover letter. 

\begin{figure}
	\centering
	
	\begin{tikzpicture}[shorten >=1pt, auto, node distance=3cm, thick, scale=0.8, every node/.style={scale=0.8}]
	
	\tikzstyle{node_style} = [circle,draw=black]
	%\tikzstyle{edge_style} = [draw=black, line width=2, ultra thick]
	\node[node_style] (gnd) at (0,0) {Gnd};
	\node[node_style] (dpt) at (0,-2) {Dpt};
	\node[node_style] (job) at (2,0) {Job};
	\node[node_style] (mrk) at (2,-3) {Mrk};
	\node[node_style] (age) at (4,1) {Age};
	\node[node_style] (cvr) at (4,-3) {Cvr};
	\node[circle,draw=black,dashed] (y) at (6,-1) {$\predY$};
	
	\draw[->]  (gnd) edge (job);
	\draw[->]  (dpt) edge (mrk);
	\draw[->]  (age) edge (job);
	
	\draw[->]  (cvr) edge (y);
	\draw[->]  (job) edge (y);
	\draw[->]  (dpt) edge (y);
	\draw[->]  (mrk) edge (y);
	
	\end{tikzpicture}
	
	\caption{Graph of the pooled model $\modeldiagram{\Mpool}$ after the first step of the \emph{pooling-removal} algorithm. \label{fig:Pool_Filt}}
\end{figure}

This example clearly illustrates the different effects of the two proposed algorithms. In the  \emph{removal-pooling} algorithm all causal connections suggested by the experts spread discrimination; the decision of a single expert (such as Alice adding a causal edge between gender and department) is sufficient to remove a whole set of nodes from the final aggregated model $\Mpool$. On the contrary, in the \emph{pooling-removal} algorithm models are first aggregated, thus removing causal edges that are not widely supported (such as the edge between gender and department filtered out by the majority pooling), and then sensitive nodes are removed.
Notice that, for both algorithms, the qualitative generation of the pooled model $\Mpool$ might be followed by the quantitative step of causal model aggregation in which the structural equations of the pooled model $\Mpool$ are computed using a pooling function \citep{bradley2014aggregating}. 

\section{Conclusion \label{sec:Conclusion}}
This paper offers a first exploration of the problem of performing aggregation of causal models under the requirement of counterfactual fairness. In particular, we explored how the two-stage approach for casual model aggregation may be adapted in its first stage to take into account the requirement of counterfactual fairness. We presented two simple algorithms, built around the idea of \emph{removal} and \emph{pooling}, to solve the problem of aggregation and we showed how they can lead to different, yet reasonable, solutions.
 
Working with causal graphs and aggregating models produced by different experts while, at the same, respecting a principle of counterfactual fairness, constitutes a relevant problem in the field of machine learning and decision making. This work may be seen as a starting point for further research  and we suggest some avenues for future development that we are investigating:

\begin{itemize}
	\item Algorithms for aggregation may be refined; the current algorithms take an extremely safe stance and discard a lot of information in order to guarantee counterfactual fairness. More subtle algorithms, working both on the first and second stage of causal model aggregation may be developed.
	
	\item The definition of protected attributes may be enriched with the introduction of \emph{resolving variables} (i.e., variables that stop the propagation of discrimination from the protected attributes) and \emph{proxy variables} (i.e., variables that propagate discrimination from the protected attributes) \cite{kilbertus2017avoiding}. %In turn, this enrichment may allow for the definition of more refined algorithms. for causal model aggregation under counterfactual fairness.
	
	\item Interesting information for compensating unfair biases may be extracted from the difference between the %potentially unfair 
	models provided by the experts and the fair aggregated model of the decision maker. Indeed, experts are likely to provide \emph{descriptive} models of the dynamics of a given system, while the decision maker is interested in coming up with a \emph{prescriptive} model of ideal behavior. The distance between the experts' models and the decision maker's may provide a measure of how close a social system is to behaving in accordance with a principle of fairness, and the specific differences may highlight variables on which to act to reduce discrimination.
	
	\item Finally, a different scenario may be considered, in which the experts are actually aware of fairness issues and %they themselves 
	provide fair models. From the point of view of the decision maker, it would be interesting to prove formally if there is an aggregation function which preserves fairness. %and whether this property may be proved formally.
\end{itemize}

\renewcommand\bibname{References}
\bibliographystyle{plainnat}
\bibliography{opinionpooling}

\end{document}